\documentclass[times,twocolumn,final]{elsarticle}

\usepackage{multirow,multicol} %fjs
\usepackage{medima}
\usepackage{framed,multirow}
\usepackage{amssymb}
\usepackage{latexsym}
\usepackage{url}
\usepackage{xcolor}
\usepackage{hyperref}
\usepackage{amsmath,amssymb,amsfonts}
\usepackage{bm}%fjs

\definecolor{newcolor}{rgb}{.8,.349,.1}
\journal{Medical Image Analysis}

\begin{document}
\verso{Fang \textit{et~al.}}
\begin{frontmatter}

\title{Deep Triplet Hashing Network for Case-based Medical Image Retrieval}

\author[1,2,4]{Jiansheng \snm{Fang}}
\author[3]{Huazhu \snm{Fu}}
\author[2]{Jiang \snm{Liu} \corref{cor}}
\cortext[cor]{Corresponding author} \ead{liuj@sustech.edu.cn}

\address[1]{School of Computer Science and Technology, Harbin Institute of Technology, Harbin 150001, China}
\address[2]{Department of Computer Science and Engineering, Southern University of Science and Technology, Shenzhen 518055, China}
\address[3]{Inception Institute of Artificial Intelligence, Abu Dhabi, United Arab Emirates}
\address[4]{CVTE Research, Guangzhou 510530, China}

\received{18 May 2020}
\finalform{10 Aug 2020}
\accepted{12 Oct 2020}
\availableonline{28 Jan 2021}
\communicated{J.Liu}

\begin{abstract}
Deep hashing methods have been shown to be the most efficient approximate nearest neighbor search techniques for large-scale image retrieval. However, existing deep hashing methods have a poor small-sample ranking performance for case-based medical image retrieval. The top-ranked images in the returned query results may be as a different class than the query image. This ranking problem is caused by classification, regions of interest (ROI), and small-sample information loss in the hashing space. To address the ranking problem, we propose an end-to-end framework, called Attention-based Triplet Hashing (ATH) network, to learn low-dimensional hash codes that preserve the classification, ROI, and small-sample information. We embed a spatial-attention module into the network structure of our ATH to focus on ROI information. The spatial-attention module aggregates the spatial information of feature maps by utilizing max-pooling, element-wise maximum, and element-wise mean operations jointly along the channel axis. To highlight the essential role of classification in diﬀerentiating case-based medical images, we propose a novel triplet cross-entropy loss to achieve maximal class-separability and maximal hash code-discriminability simultaneously during model training. The triplet cross-entropy loss can help to map the classification information of images and similarity between images into the hash codes. Moreover, by adopting triplet labels during model training, we can utilize the small-sample information fully to alleviate the imbalanced-sample problem. Extensive experiments on two case-based medical datasets demonstrate that our proposed ATH can further improve the retrieval performance compared to the state-of-the-art deep hashing methods and boost the ranking performance for small samples. Compared to the other loss methods, the triplet cross-entropy loss can enhance the classification performance and hash code-discriminability.
\end{abstract}

\begin{keyword}
\MSC 41A05\sep 41A10\sep 65D05\sep 65D17
%% Keywords
\KWD Medical Image Retrieval\sep Deep Hashing Methods\sep Spatial Attention\sep Region of Interest\sep Triplet Labels
\end{keyword}

\end{frontmatter}
%\linenumbers

\section{Introduction}
With the rapid growth of medical images produced by various radiological imaging techniques, significant attention has been devoted to the application of medical image processing technology. In recent decades, in particular, motivated by pattern recognition and computer vision techniques, such as deep learning methods, medical image processing has played an increasingly important role in assisting the diagnosis and assessment of diseases~\citep{litjens2017survey,Fu2020,Orlando2020}. However, the objective interpretation of medical images is fraught with high inter-observer variability and limited reproducibility. Further, although the outputs given by medical image classification tasks are only meant to be complementary to clinical decision-making \citep{doi2007computer}, they still inevitably affect the expert decisions when discrepancies occur. To circumvent any discrepancies between expert interpretations, Content-Based Image Retrieval (CBIR) can present prior cases with similar disease manifestations to provide a reference-based assessment. CBIR aims to produce a fine-grained ranking of a large number of candidates according to their relevance to the query by indexing and mining large image databases \citep{zhou2017recent}. In effect, this helps create a context similar to the query, thus assisting in evidence-based clinical decision-making. 

For better assistance in assessment, CBIR should have access to plenty of cases, which requires the retrieval algorithm to be both scalable and accurate. Hashing methods for CBIR have arisen as a promising solution for this, mapping high-dimensional feature descriptors to compact hash codes \citep{conjeti2017deep,wu2019deep}. The low-dimensional hash codes can preserve the semantic structure of the high-dimensional feature descriptors and are suitable for efficient data storage and fast searching \citep{zhuang2016fast}. Hashing methods can be roughly categorized into data-dependent and data-independent methods \citep{wang2017survey}. Data-independent methods focus on using random projections to construct random hash functions. Compared with the data-dependent methods, data-independent methods need longer codes to achieve satisfactory performance \citep{gong2012iterative}. Recent research focus has shifted to data-dependent methods, which learn hash functions in either a two-stage or end-to-end manner. The two-stage manner generates a vector of hand-crafted descriptors followed by learning the hashing functions. The similarity between two independent stages might not be optimally preserved by the hand-crafted features, and thus the learned hash codes are sub-optimal \citep{lai2015simultaneous}. To address the drawbacks of the two-stage manner, deep hashing methods \citep{li2015feature} with end-to-end training have been proposed to simultaneously learn image features and hash codes with deep neural networks, and have demonstrated superior performance over traditional hashing methods \citep{zheng2017sift}.

\begin{figure}[htb]
  \centering
  \includegraphics[width=\linewidth]{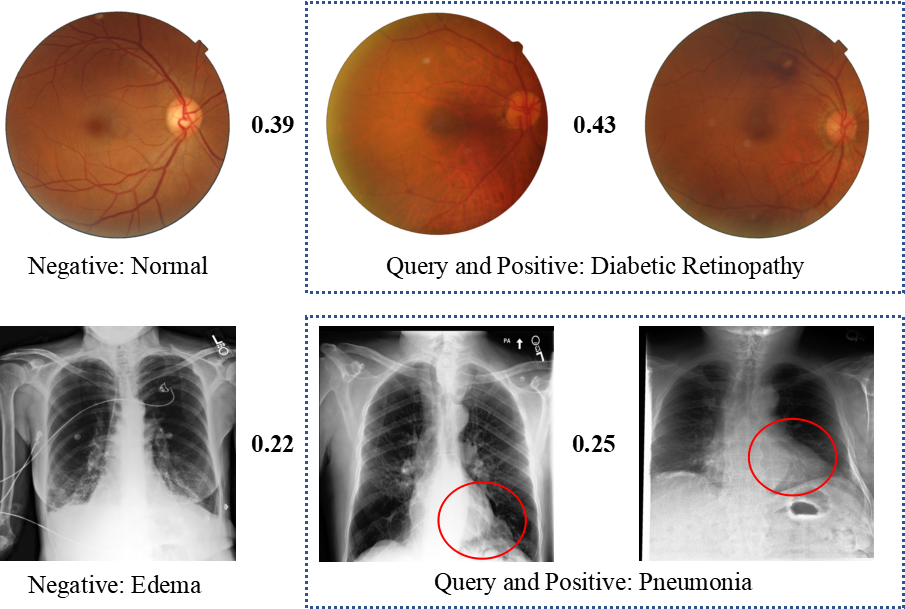}
  \caption{Schematic of ranking problem. Each row represents a triplet sample, where the query image and positive image enclosed by a rectangle are from the same class. Each image is represented by a 36-bit hash code generated by existing deep hashing methods. Here we can see that the distance of hash codes between the query image and positive image is greater than the query image and negative image.}
  \label{fig1}
\end{figure} 
Currently, deep hashing methods are widely applied for application-specific medical image retrieval, such as deep multi-instance hashing for tumor assessment \citep{conjeti2017deep}, deep residual hashing for chest X-ray images \citep{conjeti2017hashing}, order-sensitive deep hashing for multi-morbidity medical image retrieval \citep{chen2018order}, etc. However, the hash codes learned by existing hashing methods usually lose information related to classification, regions of interest (ROI), and small samples. These three pieces of information play an essential role in enhancing the ranking quality of small samples in case-based medical image retrieval. As shown in Fig.~\ref{fig1}, the negative image is more similar to the query image than the positive image in the hashing space. Such a ranking problem originates from the fact that information relating to classification, ROIs, and small samples is not fully mapped into the hash codes. For example, a very prominent wedge-shaped airspace consolidation in the left lung (red circle) is so small that its information embedded into the compact hash codes is drown out in the whole X-ray image. Clinically, the manifestation of the wedge-shaped region is characteristic of bacterial pneumonia. If the information of this region could be fully learned and mapped in the hash codes, this region could be used to discriminate between pneumonia and edema. Moreover, the information regarding small samples and their classes should be fully utilized during model training, so that the corresponding information can be mapped into the hash codes to play essential roles in differentiating case-based medical images.

To address the ranking problem, in this work, we present an end-to-end deep triplet hashing framework to learn hash codes with maximal discriminative capability, which we call the Attention-based Triplet Hashing (ATH) network. Inspired by the attention mechanism \citep{vaswani2017attention,woo2018cbam,li2019attention}, we embed a spatial-attention module into the convolutional neural network (CNN) structure to promote discriminative capability by capturing the ROI information and mapping it into the hashing space. To alleviate the imbalanced-sample problem, we utilize the information of small samples fully, with the help of triplet labels during model training. A novel triplet cross-entropy loss is proposed to preserve the classification and similarity information in the hashing space, simultaneously. Given triplet labels, with the help of the spatial-attention module and triplet cross-entropy loss, the information loss related to classification, ROIs, and small samples can be alleviated to enhance the ranking quality for case-based medical image retrieval. The main contributions of this work are summarized as follows:
\begin{itemize}
    \item[1)] An end-to-end framework, named the Attention-based Triplet Hashing (ATH) network, is presented to address the ranking problem for case-based medical image retrieval. Our ATH aims to promote the discriminative capability of hash codes by preserving the information related to classification, ROIs, and small samples in the hashing space.
    \item[2)] A spatial-attention module is embedded into the ATH network to boost the ROI representation by focusing on ROI information in the whole medical image. With the help of a novel triplet cross-entropy loss, maximal class-separability and maximal hash code-discriminability are simultaneously achieved during model training. To alleviate the imbalanced-sample problem to some extent, the hash codes are learned with the help of triplet labels in order to fully utilize small samples.
    \item[3)] Extensive experiments on two case-based medical datasets demonstrate that our proposed ATH can further improve the retrieval performance compared to the state-of-the-art deep hashing methods and boost the ranking performance for small samples. Compared to the other loss methods, the triplet cross-entropy loss can enhance the classification performance and hash code-discriminability. Our code and model have been released in \url{https://github.com/fjssharpsword/ATH}.
\end{itemize}
The rest of this paper is organized as follows: Section~\ref{sec_related} discusses related works. Section~\ref{sec_method} describes our methodology in detail. Section~\ref{sec_exp}  extensively evaluates the proposed method on two medical images datasets. Section~\ref{sec_con} gives concluding remarks.

\section{Related Works}
\label{sec_related}
Existing hashing methods can be categorized into data-independent methods and data-dependent methods. The representative data-independent methods include Locality Sensitive Hashing (LSH) \citep{slaney2008locality} and Shift-Invariant Kernels Hashing (SIKH) \citep{raginsky2009locality}. The data-dependent methods, also called learning-based hashing methods, can be further categorized into \citep{chen2018order}: \textbf{(1)} shallow learning-based hashing methods, like Metric Hashing Forests (MHF) \citep{conjeti2016metric}, Kernel Sensitive Hashing (KSH) \citep{liu2016deep}, and Spectral Hashing (SH) \citep{weiss2009spectral}; \textbf{(2)} deep learning-based hashing methods, like Convolutional Neural Network Hashing (CNNH) \citep{xia2014supervised}, Deep Hashing (DH) \citep{erin2015deep}, Deep Hashing Network (DHN) \citep{zhu2016deep}, Simultaneous Feature Learning and Hashing (SFLH) \citep{lai2015simultaneous}, Deep Semantic Ranking based Hashing (DSRH) \citep{zhao2015deep}, and Deep Similarity Comparison Hashing (DSCH) \citep{zhang2015bit}. The former learn hashing functions in a two-stage manner from hand-crafted features such as SIFT \citep{lowe2004distinctive}, GIST \citep{oliva2001modeling}, and the hash codes learning procedure is independent of the image features, which may lead to sub-optimal performance. In contrast to the former, the latter directly tailor features for hashing in an end-to-end manner with a powerful CNN, and have shown great promise recently.

Deep hashing methods leverage ground-truth labels to preserve similarity in the hash codes. Typically, the labeled data for ranking tasks are provided in one of two forms: pairwise labels or triplet labels \citep{wang2016deep}. The canonical examples of pairwise labels are Deep Residual Hashing (DRH) \citep{conjeti2017hashing} and Deep Pairwise-Supervised Hashing (DPSH) \citep{li2015feature}. DPSH was the first deep hashing method to simultaneously perform image feature learning and hash code learning with pairwise labels, and achieves the highest performance compared to other deep hashing methods. DRH was designed to preserve similarity and generate compact hashing code by defining a similarity matrix with pairwise labels for medical image retrieval. Representative methods of triplet labels include Deep Supervised Hashing (DSH) \citep{wang2016deep} and Deep Binary Embedding Networks (DBEN) \citep{zhuang2016fast}. Because the triplet labels inherently contain richer information than pairwise labels, the DSH outperforms DPSH and other deep hashing methods. 

While the aforementioned deep hashing methods have certainly achieved some degree of success, there still exists a ranking problem in the field of case-based medical image retrieval. One of the reasons for this is the classification information loss during model training. Inspired by the circle loss, which combines the triplet loss and cross-entropy loss \citep{sun2020circle}, we propose the triplet cross-entropy loss to preserve the classification information. In the circle loss, each similarity score is given different penalties according to its distance to the optimal effect. We argue that the optimal effect is still learned from samples during model training, so the triplet cross-entropy keeps the original form of both the triplet loss and cross-entropy loss. The other reason is the small-sample information loss during model training. We utilize triplet labels for model training to overcome the imbalanced-sample problem. Each triplet label can be naturally decomposed into two pairwise labels. In the hashing space, the query image is simultaneously close to the positive image and far from the negative image. Triplet labels explicitly provide a notion of relative similarities between images, while pairwise labels can only encode this implicitly \citep{wang2016deep}. Small samples are not only used as positive images themselves but also as negative images of large samples. Thus we argue that triplet labels, which can better exploit small-sample information during model training, can help to alleviate the imbalanced-sample problem.

Similar to DSH, we propose ATH to perform image feature learning and hash code learning simultaneously by maximizing the likelihood of the given triplet labels. With the given triplet labels, our ATH enhances the ranking quality for small samples by fully utilizing the small-sample information. Different from DSH, with the help of a novel triplet cross-entropy loss, maximal class-separability and maximal hash code-discriminability are simultaneously achieved during model training. We argue that class-separability information may be lost when punishing the similarity loss of global features. By conducting validations on chest X-ray images, DRH was demonstrated its better performance by preserving the class-separability. In DRH, a retrieval loss inspired by neighborhood component analysis is used for learning discriminative hash codes. However, the retrieval loss in DRH is only suitable for the co-occurring manifestation of multiple diseases. In this work, the policy on preserving the class-separability is to directly punish the classification loss and similarity loss simultaneously. 

In addition to the triplet cross-entropy loss, a spatial-attention mechanism \citep{zhu2019empirical} is also introduced for case-based medical retrieval. Recently, attention mechanisms have been successfully applied in CNNs, significantly boosting the performance of many medical image tasks \citep{oktay2018attention,nie2018asdnet}, including segmentation, recognition, and classification. For instance, an attention-based CNN \citep{li2019attention} is proposed for glaucoma detection, including an attention prediction subnet, a pathological area localization subnet, and a glaucoma classification subnet. A novel Attention Gate (AG) \citep{schlemper2019attention} can also be easily integrated into standard CNN models to leverage salient regions in medical images for various medical image analysis tasks, including fetal ultrasound classification, and 3D CT abdominal segmentation. Attention mechanisms improve the performance by guiding the model activations to be focused around salient regions. Based on prior research, we argue that the spatial-attention mechanism can also be beneficial to the performance of medical image retrieval by capturing the ROI information. Different from the average-pooling and max-pooling applied in CBAM \citep{woo2018cbam}, we utilize max-pooling, element-wise maximum, and element-wise mean operations jointly along the channel axis to generate an efficient feature descriptor.

Based on the above discussion related to the novelty of this work, the proposed ATH has two key components: \textbf{(1)} a medical image feature learning component with a spatial-attention module; \textbf{(2)} a hash code learning component for image features, with the triplet cross-entropy loss. Extensive experiments on two medical image datasets demonstrate the effectiveness of our ATH.

\section{Proposed Methodology}
\label{sec_method}

Our ATH aims to learn compact hash codes from original medical images with the given triplet labels. The hash codes should meet three requirements. (a) The query image should be encoded close to positive images and far from negative images in the hashing space. (b) The ROI information should be effectively encoded in discriminative hash codes. (c) The information related to small samples and their class should be fully mapped into the hash codes. Based on the spatial-attention mechanism and triplet cross-entropy loss, ATH is trained in an end-to-end manner, in which image feature learning and hash code learning from triplet labels are performed simultaneously.

\subsection{Attention-based Network Architecture}
Analytically, for the task of case-based medical image retrieval, the small-sample ranking problem originates from classification, ROI, and small-sample information loss in the hash codes. The fundamental goal of medical image retrieval tasks is to present prior cases with similar disease manifestations to assist evidence-based clinical decision-making. However, if the prior cases with similar disease manifestations (same class) rank lower than expected in the returned query list, the ranking quality not only impacts the effectiveness of clinical decision-making but also likely leads to error-prone diagnosis. To improve the ranking quality, we propose corresponding solutions for the classification, ROI, and small-sample information loss, including a novel triplet cross-entropy loss, a spatial-attention mechanism, and triplet labels.

\begin{figure*}[htb]
  \centering
  \includegraphics[width=\linewidth]{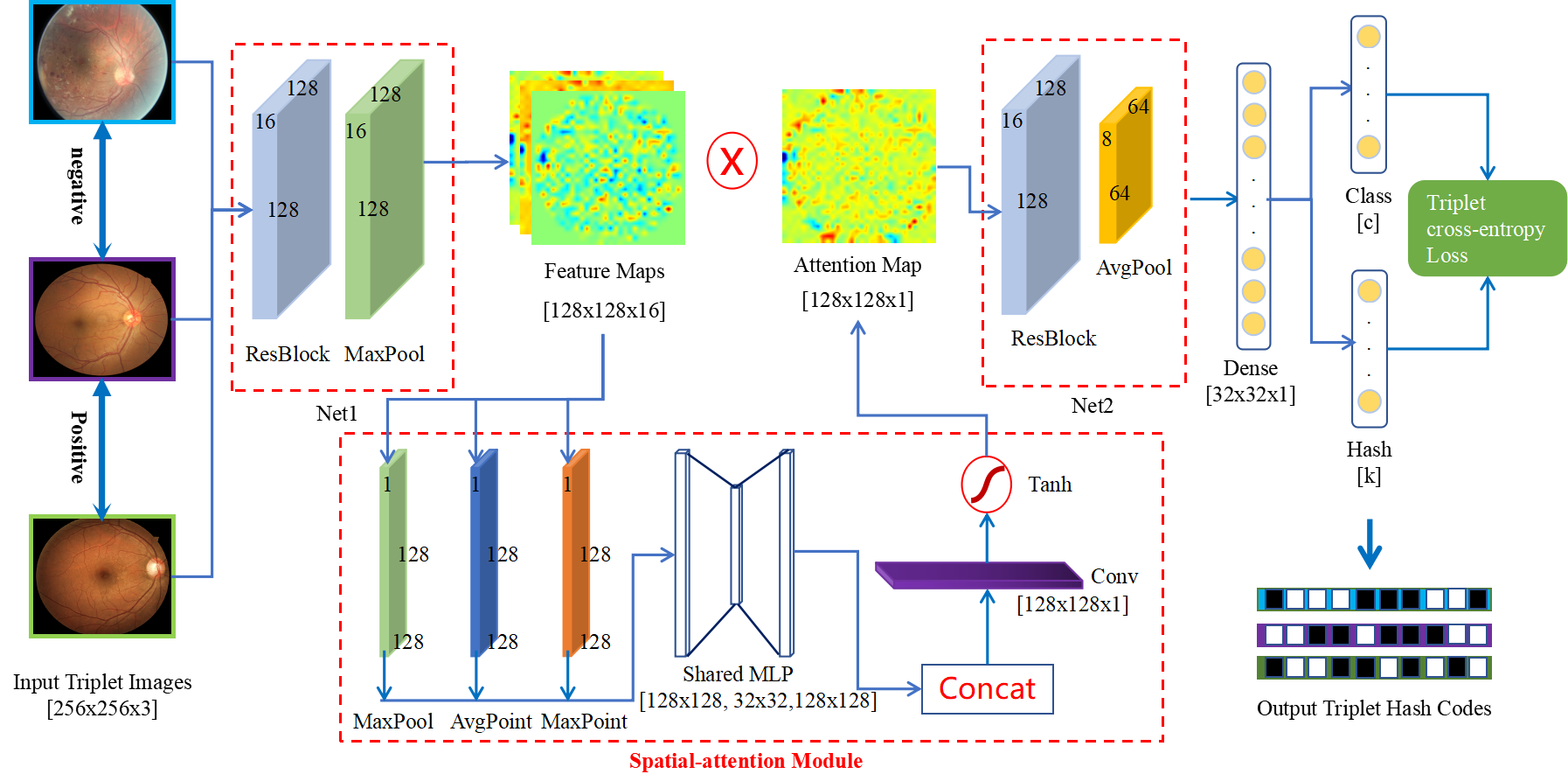}
  \caption{Illustration of our ATH network structure. A spatial-attention module aggregates spatial information of feature maps by utilizing max-pooling (MaxPool), element-wise maximum (MaxPoint), and element-wise mean (AvgPoint) operations jointly along the channel axis. A dense layer is simultaneously mapped into the hash code-generation layer and classification output layer for training the triplet cross-entropy loss. The input triplet images share the network weights and are mapped into the corresponding triplet hash codes in which the classification, ROI, small-sample information is fully embedded.}
  \label{fig2}
\end{figure*} 
As shown in Fig.~\ref{fig2}, we present an attention-based triplet hashing network to jointly learn visual feature extraction and the subsequent mapping to compact hash codes. The architecture of our ATH consists of a net1 module, a spatial-attention module, and a net2 module, and terminates in a dense layer for hash code-generation and classification outputs. The net1 module contains a residual block \citep{he2016deep} and a max-pooling layer followed by a spatial-attention module. The spatial-attention module generates an attention map that is multiplied with the input feature maps. After the net2 module, which contains a residual block and an average-pooling layer, the dense layer is designed as a convolutional layer with \begin{math}32 \times 32 \times 1\end{math} nodes according to class activation mapping (CAM) \citep{zhou2016learning}. The dense layer is separately mapped into the hash code-generation layer with \begin{math}k\end{math} nodes and the classification output layer with \begin{math}c\end{math} nodes. All the convolutional and pooling layers use \begin{math}3 \times 3\end{math} filters with stride 2 and are followed by batch normalization \citep{ioffe2015batch}. All the convolutional layers and the fully connected layer are equipped with the ReLU \citep{nair2010rectified} activation function. The triplet images are input into the ASH at the same time to generate triplet hash codes and share network weights during training. 

In ATH, the spatial-attention module inputs the feature maps \begin{math}\bm{F}\end{math} with \begin{math}128 \times 128 \times 16\end{math} dimensions and outputs an attention map \begin{math}\bm{M(F)}\end{math} with \begin{math}128 \times 128 \times 1\end{math} dimensions by utilizing the inter-spatial relationship of features. We argue that each ROI region mainly consists of informative parts and salient points, both of which should respond to the gradient back-propagation. To focus on salient points, we use element-wise maximum (MaxPoint) and element-wise mean (AvgPoint) operations along the channel axis to generate two different spatial context descriptors: \begin{math}\bm{F}_{avg}\end{math} and \begin{math}\bm{F}_{max}\end{math}. The element-wise maximum and element-wise mean operations compute the maximum and average of each element along the channel axis, respectively. \begin{math}\bm{F}_{avg}\end{math} and \begin{math}\bm{F}_{max}\end{math} are calculated as:
\begin{equation}
\begin{aligned} 
\bm{F}_{max} & = [f_{1}...f_{i}...f_{128 \times 128}] && f_{i}= \max \limits_{1\leq c \leq 16} \mathcal{X}_{i}(c) \\
\bm{F}_{avg} & = [f_{1}...f_{i}...f_{128 \times 128}] && f_{i}= \overline{\mathcal{X}_{i}(c)}
\end{aligned}
\label{eq1}, 
\end{equation}
where \begin{math}\mathcal{X}_{i}(c)\end{math} is the response of the \begin{math}i^{th}\end{math} point in \begin{math}c^{th}\end{math} channel. To focus on informative parts, we use a max-pooling (MaxPool) operation to generate a max-pooled feature descriptor: \begin{math}\bm{F}_{maxp}\end{math}. All three descriptors are independently fed forward to a shared multi-layer perceptron (MLP) for denoising. Inspired by maximum activation of convolutions~\citep{tolias2015particular}, the MaxPoint operation encodes the maximal point response across feature maps of the convolutional layers. It is different from the MaxPool operation which encodes the maximal local response of each of the convolutional layers. The shared MLP contains a hidden layer, and the hidden activation size is set to the size of the dense layer. The MLP weights are shared by the three input descriptors and followed by the ReLU activation function. After the MLP, we concatenate and convolve the three descriptors into an attention map that encodes which areas to emphasize or suppress. The convolutional layer has a filter size of 3x3 and is followed by the tangent activation function. In short, the spatial-attention is computed as:
\begin{equation}
\bm{M(F)} = \mathcal{T} (\mathcal{F}([MLP(\bm{F}_{avg}); MLP(\bm{F}_{max}); MLP(\bm{F}_{maxp})] ) )
\label{eq2},
\end{equation}
where \begin{math}\mathcal{T}\end{math} denotes the tangent function, \begin{math}\mathcal{F}\end{math} represents a convolution operation, and \begin{math}MLP\end{math} is the operator of the shared MLP.

Based on the feature maps generated by the net1 module, we apply max-pooling, element-wise maximum, and element-wise mean operations jointly along the channel axis to generate a three-channel feature map. Each operation gathers an important clue related to the ROI information to infer a finer spatial-attention by computing spatial statistics. For each pixel, the maximum and average along the channel axis are computed for describing the channel-wise context features. The max-pooling operation prevents the marginal value of the ROI region from weakening. By exploiting both element-wise maximum and element-wise mean outputs independently, the features with distinct levels are input into the shared MLP module for further denoising. Such a spatial-attention module is beneficial for capturing the ROI information in high-resolution medical image. Using an attention-based CNN enables the learning procedure of our ATH to focus on ROI feature extraction on the raw pixels of input images. The hierarchical non-linear function exhibits a powerful learning capacity and encourages the learned feature to capture the ROI information by using the spatial-attention module.

\subsection{Triplet Cross-entropy Loss}
To overcome the ranking problem shown in Fig.~\ref{fig1}, we propose a novel triplet cross-entropy loss to achieve maximal-class separability and maximal hash code-discriminability by punishing similarity and classification losses simultaneously. The main idea behind the triplet cross-entropy loss is that the similarity and classification information should be simultaneously preserved during model training from triplet input images.

Mathematically, given \begin{math}m\end{math} training images \begin{math}\bm{I}=\{I_{1},\dots,I_{m}\}\end{math} and class labels \begin{math}\bm{L}=\{1,\dots,c\}\end{math}, the triplet labels \begin{math}\bm{T}=\{(Q_{1},P_{1},N_{1}),\dots,(Q_{i},P_{i},N_{i}),\dots,(Q_{m},P_{m},N_{m})\}\end{math} are generated by randomly selecting two images as a query image and a positive image from the same class (\begin{math}Q_{i}\end{math} and \begin{math}P_{i}\end{math}) and randomly selecting a negative image from different classes (\begin{math}Q_{i}\end{math} and \begin{math}N_{i}\end{math}). In the triplet labels, the query image of index \begin{math}Q_{i}\end{math} is similar to the positive image \begin{math}P_{i}\end{math} and dissimilar to the negative image \begin{math}N_{i}\end{math}, where the index \begin{math}i \in \{1,\dots,m\}\end{math} is randomly selected from the \begin{math}m\end{math} training images. When sampling triplet labels, small samples are selected as the negative image of large samples. In other words, small samples are reused multiple times during model training to preserve the information in the hash codes.

Generally speaking, ATH aims at learning a mapping from input triplet images to triplet hash codes \begin{math}\bm{H}_{Q}=\{\bm{h_{1,k}},\dots,\bm{h_{m,k}}\}\end{math}, \begin{math}\bm{H}_{P}=\{\bm{h_{1,k}},\dots,\bm{h_{m,k}}\}\end{math}, and \begin{math}\bm{H}_{N}=\{\bm{h_{1,k}},\dots,\bm{h_{m,k}}\}\end{math}. For scalable retrieval, the length of hash code \begin{math}k\end{math} is much smaller than the dimension of medical image. The distance between \begin{math}\bm{H}_{Q}\end{math} and \begin{math}\bm{H}_{P}\end{math} should be smaller than the distance between \begin{math}\bm{H}_{Q}\end{math} and \begin{math}\bm{H}_{N}\end{math}. More specifically, the ROI information should be effectively encoded in the hash codes, such that the distances of similar images or dissimilar images can be affected by the ROI information. Based on the design of the triplet cross-entropy loss, ATH is trained with triplet labels and ground-truth labels to perform hash code learning and classification likelihood learning simultaneously. Corresponding to the triplet labels \begin{math}\bm{T}\end{math}, the ground-truth labels are \begin{math}\bm{Y}=\{(y_{Q_{1}},y_{P_{1}},y_{N_{1}}),\dots,(y_{Q_{i}},y_{P_{i}},y_{N_{i}}),\dots,(y_{Q_{m}},y_{P_{m}},y_{N_{m}})\}\end{math}, where \begin{math}y_{Q_{i}}, y_{P_{i}}, y_{N_{i}} \in \bm{L}\end{math}.

To punish the similarity loss given the triplet labels, a simple distance (e.g. Euclidean distance) is used to compare the similarity in the target space by mapping the input to the target space. In the training phase, the codes of query images and positive images should be as close as possible, while the codes of query images and negative images should be far away. Based on this objective, a hinge ranking loss form is naturally designed to minimize the distance between similar image pairs and maximize the distance between dissimilar image pairs. The loss with respect to triplet labels is defined as:
\begin{equation}
\mathcal{L}(\bm{T}) = \max \{ r \cdot k - Dist(\bm{H}_{Q},\bm{H}_{N}) + Dist(\bm{H}_{Q},\bm{H}_{P}),0\}
\label{eq3},
\end{equation}
where \begin{math}Dist(\cdot,\cdot)\end{math} denotes the L2-norm to measure the distance between hash outputs, \begin{math}k\end{math} is the length of hash codes, and \begin{math}r\in[0,1]\end{math} is a weighting parameter that controls the punishment strength of differentiating degrees between dissimilar images. The triplet loss is applied in such a way that only dissimilar pairs with the distance between them being within a specific radius are eligible to contribute to the loss. When \begin{math}r=0\end{math}, there is no punishment for dissimilar images mapped to close hash codes. When \begin{math}r=1\end{math}, the hash codes of dissimilar images must be completely different. If \begin{math}r=0.5\end{math}, this implies that half of the hash code lengths between dissimilar images should be different. The triplet loss is designed to measure how well the given triplet labels are satisfied by the learned hash codes by computing the likelihood of the given triplet labels.

To punish the classification loss given the ground-truth labels, we define the cross-entropy loss by jointly considering the input triplet images, as follows:
\begin{equation} 
\mathcal{L}(\bm{T},\bm{Y}) = \sum_{i=1}^{m} \{CE(\hat{y}_{Q_{i}},y_{Q_{i}}) + CE(\hat{y}_{P_{i}},y_{P_{i}}) + CE(\hat{y}_{N_{i}},y_{N_{i}})\} , 
\label{eq4} 
\end{equation}
where \begin{math}CE(\cdot,\cdot)\end{math} denotes the common cross-entropy loss form, and \begin{math}\hat{y}\end{math} denotes the predicted class. With the similarity loss \begin{math}\mathcal{L}(\bm{T})\end{math} and classification loss \begin{math}\mathcal{L}(\bm{T},\bm{Y})\end{math}, we reverse the sum of both to update the weights of the model. Theoretically, the cross-entropy loss is beneficial for preserving the classification information in the hash codes, and the triplet loss can also help to improve the classification performance by encouraging hash codes to minimize intra-class similarity and maximize inter-class similarity. 

\section{Experiments}
\label{sec_exp} 

\subsection{Datasets}
\begin{itemize}
    \item[1)] \textbf{Fundus-iSee:} The private ophthalmic Fundus-iSee dataset with four disease classes consists of 10,000 high-resolution images labeled by professional doctors with rich clinical experience specifically. There are 720 images of age-related macular degeneration (AMD), 270 images of diabetic retinopathy (DR), 450 images of Glaucoma, 790 images of Myopia, and 7770 images of Normal. We randomly extract ten percent of each class for the query test, for a total of 1,000 images.
    \item[2)] \textbf{MIMIC-CXR:} The public MIMIC-CXR \citep{johnson2019mimic} dataset is a large publicly available dataset of chest radiographs that are used to identify acute and chronic cardiopulmonary conditions and to assist in related medical workups. We randomly select two groups of images with frontal view from four classes, including Normal, Edema, Pneumonia, Fracture. The training set contains 14,555 images of Normal, 1,837 images of Edema, 3,220 images of Pneumonia, 388 images of Fracture. The test set contains 14,854 images of Normal, 944 images of Edema, 3,788 images of Pneumonia, 414 images of Fracture. The ratio of the training set and test set is 1 to 1.
\end{itemize}

\subsection{Evaluation Settings}
Three metrics are adopted to measure the precision and retrieval quality in our experiments. \begin{math}(1)\end{math} \textbf{Hit Ratio (HR).} HR is designed to measure how many images in the returned list are similar to the query image. \begin{math}(2)\end{math} \textbf{Average Precision (AP).} In the returned list, AP averages the rank positions of images similar to the query image to measure the rank quality. \begin{math}(3)\end{math} \textbf{Reciprocal Rank (RR).} RR refers to the reciprocal of the ranking of the first similar image in the returned list. The two datasets selected have a serious imbalanced-sample problem from the perspective of classification tasks. To validate the effectiveness of the triplet cross-entropy loss in preserving classification information in the hash codes, we apply \textbf{Specificity} and \textbf{Sensitivity} to measure the accuracy of classification.

In our comparative study, we use four deep hashing methods to evaluate the retrieval accuracy with mean HR (mHR), mean AP (mAP), mean RR (mRR), including DPSH-pairwise \citep{li2015feature}, DRH-pairwise \citep{conjeti2017hashing}, DSH-triplet \citep{wang2016deep}, and DBEN-triplet \citep{zhuang2016fast}. Two of the comparative methods use AlexNet \citep{krizhevsky2012imagenet} as the backbone, including DPSH-pairwise and DSH-triple. Recently, the residual block \citep{he2016deep} has been used popularly as the backbone in deep hashing methods, such as DRH-pairwise and DBEN-triplet, and shows the advantage of feature extraction. In our ATH, both the net1 and net2 modules also use the residual block as the backbone. Hence, compared to DRH-pairwise and DBEN-triplet, the advantage of our ATH framework can be demonstrated. To compare the effectiveness of the triplet cross-entropy loss, we introduce different loss methods into our ATH network, including cross entropy loss (ATH-CE), focal loss (ATH-focal) \citep{lin2017focal}, circle loss (ATH-circle) \citep{sun2020circle}, pairwise loss (ATH-pairwise), and triplet loss (ATH-triplet). 

Our ATH is implemented with pyTorch, and the network structure of ATH is illustrated in Fig.~\ref{fig2}. All deep hashing methods are trained from scratch, setting the batch size as 10 and the iteration number as 50. The parameters of comparative methods are set according to their implementation details in the corresponding papers, and the best performance is reported. The input triplet images are randomly sampled in every iteration. For the triplet cross-entropy loss in our ATH, the margin threshold \begin{math}r\cdot k\end{math} should be considerably set to suit the clinical datasets better. Without loss of generality, the weighting parameter \begin{math}r\end{math} and the length of hash codes \begin{math}k\end{math} are synchronous to each other, so we empirically set \begin{math}r=0.5\end{math} in the experiments. Without a special description, we report all the performances over 36-bit hash codes and weighting parameter \begin{math}r\end{math} of 0.5, and the top-10 similar images are returned and ranked in every retrieval. Our ATH is implemented under Pytorch framework and experiments are run on Geforce RTX 2080 Ti. In our work, the indexing and similarity calculation for evaluation uses Faiss \citep{johnson2019billion} which is a library for efficient similarity search and clustering of dense vectors.

\subsection{Results and Analysis}
To enhance the ranking quality of case-based medical image retrieval, we argue that the classification, ROI, and small-sample information loss in the hashing space need to be overcome. For ROI feature learning, we embed the spatial-attention module into the network to capture the ROI information. To preserve classification information, we propose the triplet cross-entropy loss to punish the similarity and classification losses simultaneously during model training. To overcome the imbalanced-sample problem, we sample the triplet labels from classification datasets in order to fully utilize the small-sample information. 

\subsubsection{Observation of ranking quality} 

\begin{table}[!t]
\renewcommand\arraystretch{1.2}
\caption{The performances (mHR, mAP, mHR) on the Fundus-iSee and MIMIC-CXR datasets}
\begin{center}
    \begin{tabular}{ |c|c|c|c|c|c| } 
    \hline 
    \textbf{Dataset} & \textbf{Methods} & \textbf{mHR} & \textbf{mAP} & \textbf{mRR} \\
        \hline 
        \multirow{9}*{Fundus-iSee}
        & DPSH-pairwise & 0.6063 & 0.5182 & 0.8077  \\
        & DRH-pairwise & 0.6761 & 0.5889 & 0.8555  \\
        & DSH-triplet & 0.6146 & 0.5291 & 0.8187  \\
        & DBEN-triplet & 0.7074 & 0.6378 & 0.8918  \\
        & ATH-CE & 0.6558 & 0.5696 & 0.8441  \\
        & ATH-focal & 0.6642 & 0.5807 & 0.8549  \\
        & ATH-circle & \underline{0.7180} & \underline{0.6621} & \underline{0.9033} \\
        & ATH-pairwise & 0.6804 & 0.5907 & 0.8688  \\
        & ATH-Triplet &  0.6981 & 0.6359 & 0.8845 \\
        & \textbf{ATH(ours)} & \textbf{0.7682} & \textbf{0.7220} & \textbf{0.9256} \\
        \hline 
        \multirow{9}*{MIMIC-CXR}
        & DPSH-pairwise & 0.7213 & 0.6810 & 0.8645  \\
        & DRH-pairwise & 0.7434 & 0.7218 & 0.9028  \\
        & DSH-triplet & 0.7699 & 0.7274 & 0.8853  \\
        & DBEN-triplet & 0.7535 & 0.7260 & 0.9156 \\
        & ATH-CE & 0.7592 & 0.7187 & 0.8943  \\
        & ATH-focal & 0.7701 & 0.7333 & 0.9049  \\
        & ATH-circle & \underline{0.7861} & \underline{0.7629} & \underline{0.9217}  \\
        & ATH-pairwise & 0.7601 & 0.7386 & 0.9078  \\
        & ATH-triplet & 0.7751 & 0.7467 & 0.9131  \\
        & \textbf{ATH(ours)} & \textbf{0.8543} & \textbf{0.8260} & \textbf{0.9668} \\
        \hline 
    \end{tabular}
\end{center}\label{tb1}
\end{table}
As shown in Table \ref{tb1}, due to the effectiveness of the triplet cross-entropy loss and the spatial-attention mechanism, our ATH can further improve the performance compared to the second-highest (underlined) deep hashing methods given the triplet labels. The higher mAP means more similar images (same class) in the returned list are ranked ahead, while the higher mHR implies that more similar images are retrieved. The better performance in terms of the mAP and mHR demonstrates that our ATH not only achieves better accuracy of retrieval but helps users to find the required disease case quickly. In terms of mAP, which is used to evaluate the ranking quality, our ATH consistently outperforms the second-highest methods (underlined) by around 8\%. From the perspective of mRR measuring ranking quality, our ATH consistently obtains gains above 0.90, which means that the first images in the returned list are nearly all from the same class as the query image. Following the convention in the literature of CBIR \citep{zhan2018instance,xiao2020deeply}, we further provide the performance of mAP over varying top-k, where k varies from 5, 10 to 20. The results of Table \ref{tb2} also confirm the benefits of our ATH. When the returned list top-k lengthens, the performance of all methods declines to some extent. We can observe that our ATH all achieves the best performance over the returned list of 5, 10, 20. According to the overall performance, we argue that our ATH can alleviate the information loss to some extent and fully map useful information into the hash codes.
\begin{table}[!t]
\renewcommand\arraystretch{1.2}
\caption{mAP over the varying number of the returned list on Fundus-iSee and MIMIC-CXR datasets}
\begin{center}
    \begin{tabular}{ |c|c|c|c|c|c| } 
    \hline 
    \textbf{Dataset} & \textbf{Methods} & \textbf{top-5} & \textbf{top-10} & \textbf{top-20}\\
        \hline 
        \multirow{9}*{Fundus-iSee}
        & DPSH-pairwise & 0.5400 & 0.5182 & 0.4946  \\
        & DRH-pairwise & 0.5970 & 0.5889 & 0.5440  \\
        & DSH-triplet & 0.5416 & 0.5291 & 0.5014  \\
        & DBEN-triplet & 0.6459 & 0.6378 & 0.6091 \\
        & ATH-CE & 0.5752 & 0.5696 & 0.4992  \\
        & ATH-focal & 0.5918 & 0.5807 & 0.5174 \\
        & ATH-circle & \underline{0.6813} & \underline{0.6621} & \underline{0.6260} \\
        & ATH-pairwise & 0.6286 & 0.5907 & 0.5262  \\
        & ATH-Triplet &  0.6696 & 0.6359 & 0.5589 \\
        & \textbf{ATH(ours)} & \textbf{0.7421} & \textbf{0.7220} & \textbf{0.6802} \\
        \hline 
        \multirow{9}*{MIMIC-CXR}
        & DPSH-pairwise & 0.7099 & 0.6810 & 0.5901  \\
        & DRH-pairwise & 0.7436 & 0.7218 & 0.6758  \\
        & DSH-triplet & 0.7595 & 0.7274 & 0.6327  \\
        & DBEN-triplet & 0.7551 & 0.7260 & 0.6960 \\
        & ATH-CE & 0.7432 & 0.7187 & 0.6397  \\
        & ATH-focal & 0.7595 & 0.7333 & 0.6862 \\
        & ATH-circle & \underline{0.7954} & \underline{0.7629} & \underline{0.7371} \\
        & ATH-pairwise & 0.7672 & 0.7386 & 0.6990  \\
        & ATH-triplet & 0.7616 & 0.7467 & 0.7137  \\
        & \textbf{ATH(ours)} & \textbf{0.8412} & \textbf{0.8260} & \textbf{0.7628} \\
        \hline 
    \end{tabular}
\end{center}\label{tb2}
\end{table}

\begin{figure*}[htb]
  \centering
  \includegraphics[width=\linewidth]{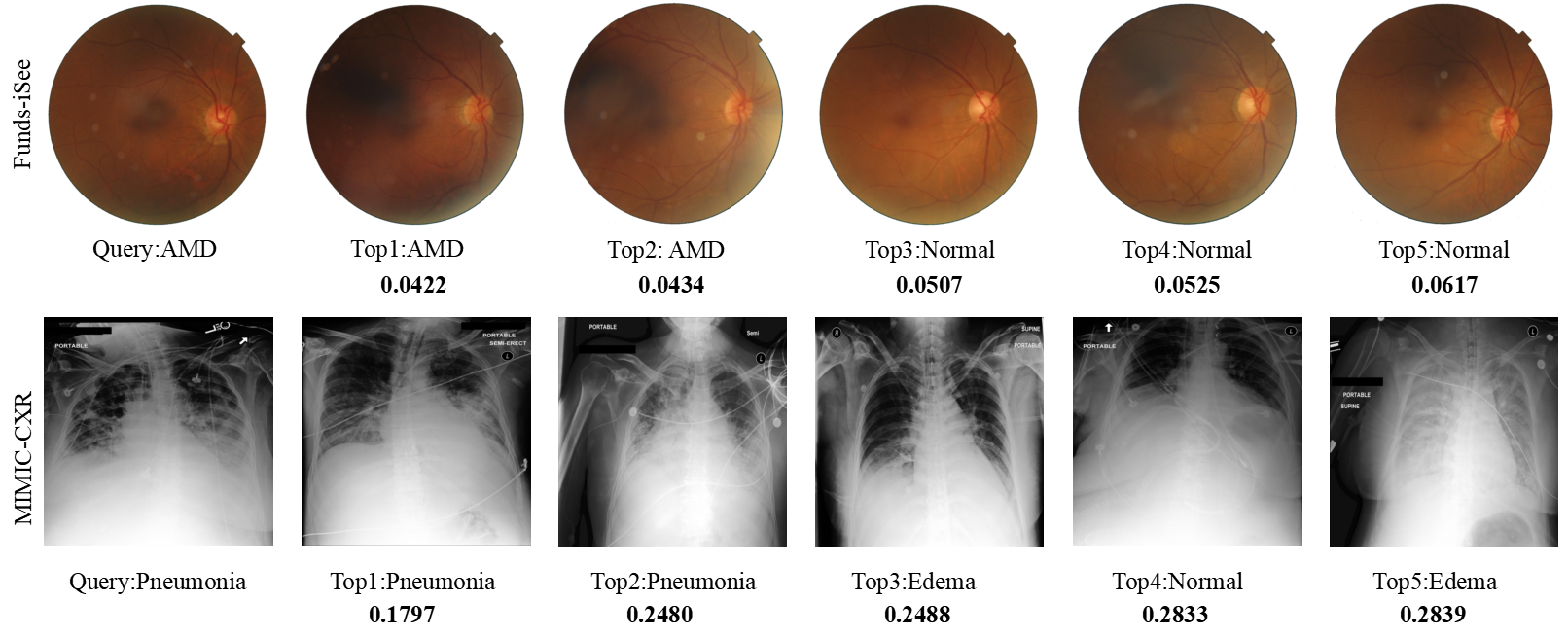}
  \caption{Qualitative results of ranking quality. Upper row: querying an age-related macular degeneration (AMD) image, top-5 similar images with hash code distance are returned in the Fundus-iSee dataset. Lower row: querying a pneumonia image, top-5 similar images with hash code distance are returned in the MIMIC-CXR dataset.}
  \label{fig3}
\end{figure*}
We can conclude two points from Table \ref{tb1} and Table \ref{tb2}. $(1)$ The two best results of both datasets are related to the network structure of ATH, in which the spatial-attention module plays a significant role. $(2)$ The order of performance (ATH$>$ATH-circle$>$ATH-triplet) demonstrates the merit of combining the triplet loss and the cross-entropy loss. Due to the triplet labels, which enable the small-sample information to be fully used during training, our triplet cross-entropy loss outperforms the circle loss. Next, we further observe the ranking quality in Fig.~\ref{fig3}. The distance between the hash codes of the same class is closer than that between different classes. By querying an AMD image, two AMD images are hit and ranked ahead in the Fundus-iSee dataset. By querying a pneumonia image, two pneumonia images are hit and ranked ahead in the MIMIC-CXR dataset. Clinically, the symptoms and manifestations of edema are similar to pneumonia. On the whole, our ATH can effectively solve the ranking problem in Fig.~\ref{fig1}.

\begin{figure*}[htb]
  \centering
  \includegraphics[width=\linewidth]{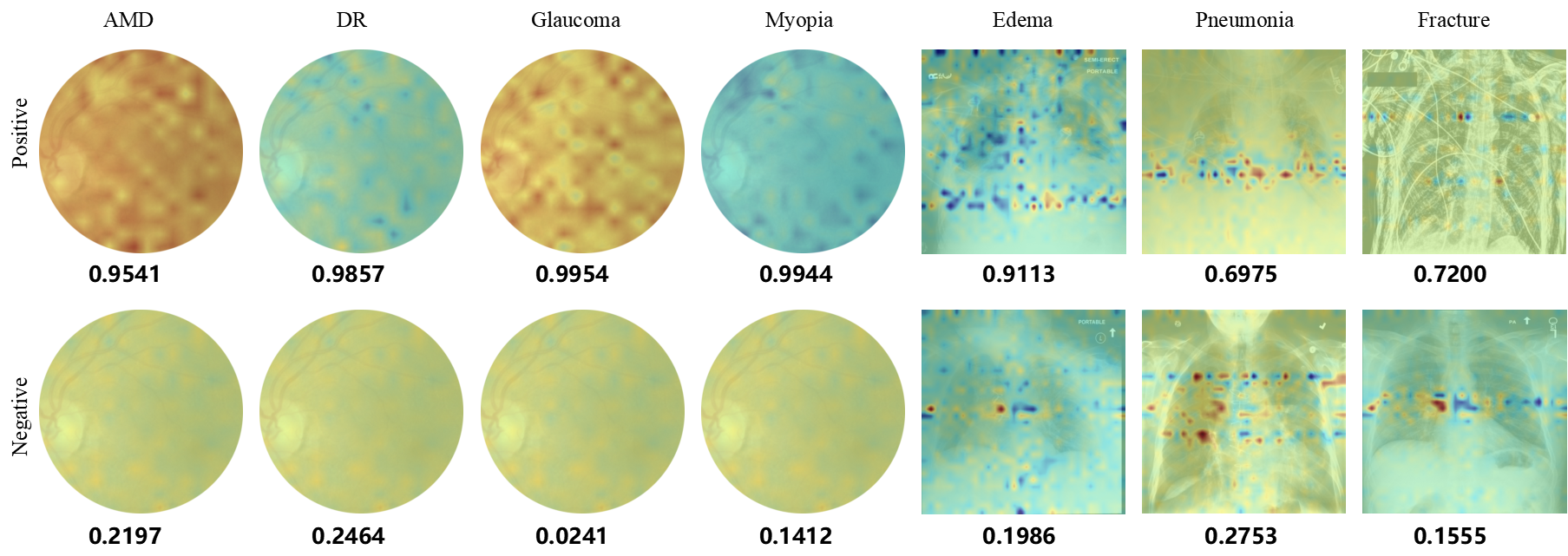}
  \caption{Qualitative results of heat maps. Heat maps of the dense layer outputs are generated according to the class activation mapping. The upper row and lower row are the positive and negative results with predicting probability, respectively. The first four classes from left to right are Age-related Macular Degeneration (AMD), Diabetic Retinopathy (DR), Glaucoma, and Myopia in the Fundus-iSee dataset, and the last three classes are Edema, Pneumonia, Fracture in the MIMIC-CXR dataset. For example, by query an AMD image, the result predicted as AMD is positive, and predicted as non-AMD is negative. The negative result belongs to AMD with a probability of 0.2197, and the positive result is 0.9541. }
  \label{fig4}
\end{figure*} 
Based on the above experimental results and analysis, we have demonstrated the effectiveness of our ATH in addressing the ranking problem by fully mapping useful information into the hash codes. To illustrate the function of the spatial-attention mechanism in capturing ROI information, we provide heat maps of the dense layer outputs in Fig.~\ref{fig4}. The heat maps of ROI regions between positive and negative results vary in degree and location. Such differences in ROI heat maps have an effect on class-separability. Each class has its distinctive ROI heat maps. Based on this observation, we argue that the features of ROI regions captured by the spatial-attention module are mapped into the hash codes. To further evaluate the contribution of the proposed spatial-attention module, we substitute the spatial-attention module with AG and CBAM in our ATH framework to compare the performance. Both AG and CBAM have been widely applied in networks and achieve competitive performance in many tasks of computer vision. For mAP of the Fundus-iSee dataset and the MIMIC-CXR dataset, our spatial-attention module averagely outperforms AG by 6.9\% and CBAM by 7.3\%. This experiment shows that the proposed spatial-attention module can achieve better performance than AG and CBAM by integrating the three spatial descriptors: \begin{math}\bm{F}_{avg}\end{math}, \begin{math}\bm{F}_{max}\end{math}, and \begin{math}\bm{F}_{maxp}\end{math}. Based on the above observation, the capability of capturing the salient value of the spatial-attention module has been attested. The spatial-attention mechanism can contribute to the hash code-discriminability by capturing the ROI information.

\subsubsection{Observation of the triplet cross-entropy loss} 
\begin{table*}[!t]
\renewcommand\arraystretch{1.2}
\caption{mAP of each class on the Fundus-iSee and MIMIC-CXR datasets.}
\begin{center}
    \begin{tabular}{|l|c|c|c|c|c|c|c|c|c|}
    \hline 
    \multirow{2}{*}{\textbf{Methods}} & \multicolumn{5}{c|}{\textbf{Fundus-iSee Dataset}} & \multicolumn{4}{c|}{\textbf{MIMIC-CXR Dataset}}\\
    \cline{2-10}
     & \textbf{Normal} & \textbf{AMD} & \textbf{DR} & \textbf{Glaucoma} & \textbf{Myopia} & \textbf{Normal} & \textbf{Edema} & \textbf{Pneumonia} & \textbf{Fracture} \\
    \hline
    DPSH-pairwise & 0.6541 & 0.0299 & 0.0018 & 0.0126 & 0.0272 & 0.5940 & 0.0126 & 0.1916 & 0.0052\\
    DRH-pairwise & 0.6533 & 0.0264 & 0.0094 & 0.0142 & 0.1422 & 0.6732 & 0.2119 & 0.7831 & 0.0099\\
    DSH-triplet & 0.6623 & 0.0236 & 0.0080 & 0.0075 & 0.0216 & 0.5906 & 0.0321 & 0.1637 & 0.0096\\
    DBEN-triplet & 0.6711 & 0.0182 & 0.0123 & 0.0023 & 0.1735 & 0.7644 & 0.2158 & 0.7989 & 0.0114 \\
    ATH-CE & 0.6629 & 0.0214 & 0.0091 & 0.0181 & 0.0550 & 0.7340 & 0.1026 & 0.5240 & 0.0147\\
    ATH-focal & 0.6865 & 0.0244 & 0.0073 & \underline{0.0272} & 0.0494 & 0.7284 & 0.0547 & 0.4373 & 0.0095 \\
    ATH-circle & 0.7279 & \underline{0.0360} & \underline{0.0142} & 0.0136 & 0.0491 & \underline{0.7719} & 0.1035 & \underline{0.8368} & 0.0087 \\
    ATH-pairwise & 0.6935 & 0.0328 & 0.0089 & 0.0122 & 0.0375 & 0.7388 & 0.0732 & 0.8123 & \underline{0.0182} \\
    ATH-triplet & \underline{0.7734} & 0.0168 & 0.0028 & 0.0062 & \underline{0.3384} & 0.7520 & \underline{0.2771} & 0.7656 & 0.0038 \\
    \textbf{ATH(ours)} & \textbf{0.8564} & \textbf{0.0425} & \textbf{0.0185} & \textbf{0.0294} & \textbf{0.5995} & \textbf{0.8220} & \textbf{0.3068} & \textbf{0.8571} & \textbf{0.0332}\\
    \hline
    \end{tabular}
\end{center}\label{tb3}
\end{table*}
The triplet cross-entropy loss for model training achieves maximal class-separability and maximal hash code-discriminability simultaneously. On the one hand, as shown in Table \ref{tb3}, our ATH achieves the best performance of the mAP over each class on the two datasets. In the ablation study of the loss function, including ATH-focal, ATH-circle, ATH-pairwise, and ATH-triplet, all of them can achieve second-highest performance on some class (underlined). Our ATH can improve the ranking quality of small samples by mapping each class's semantic information into the hash codes without sacrificing the performance of large samples. On the other hand, with the help of the triplet cross-entropy loss, we can get the classification results for the task of medical image retrieval. As shown in Table \ref{tb4}, the sensitivity performance of small samples shows that the triplet cross-entropy loss is superior to the focal loss and circle loss (underlined) without affecting the performance on the large samples. Recently, the focal loss and circle loss have demonstrated their superiority in alleviating the imbalanced-sample problem. Compared to the focal loss and circle loss, the triplet cross-entropy loss achieves maximal class-separability and maximal hash code-discriminability simultaneously during model training. 

\begin{figure}[htb]
  \centering
  \includegraphics[width=\linewidth]{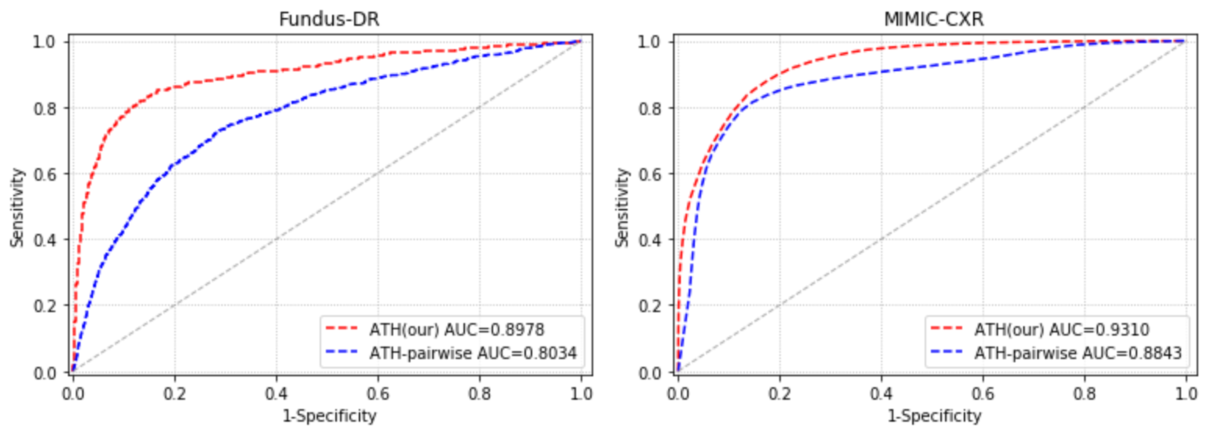}
  \caption{The ROC curves with AUC scores of our ATH (red line) and ATH-pairwise (blue line) on the Fundus-iSee and MIMIC-CXR datasets.}
  \label{fig5}
\end{figure}
For case-based medical image retrieval, a common issue is the imbalanced-sample due to the scarce disease cases. Fewer cases of certain types of diseases lead to their low sensitivity and high missed diagnosis rate. To alleviating the imbalanced-sample problem, triplet labels are input into the model to increase the use of the small-sample information. Compared to the pairwise labels, the sampling mechanism of triplet labels demonstrates advantages in fully using small samples, as shown by observing the Receiver Operating Characteristic (ROC) curves with Area-Under-the-Curve (AUC) in Fig.~\ref{fig5}. We argue that the triplet labels can play a role in maximizing inter-class distance and minimizing intra-class distance by utilizing small-sample information fully. The highest accuracy of classification means that the feature in the ATH network fully contains the information related to small samples and their classes. The classification output layer and the hash code-generation layer share the feature layers in the ATH network and are generated with the dense layer. Thus, given triplet labels, we can argue that the classification information can be mapped into the hash codes by using the triplet cross-entropy loss and can help to improve the hash code-discriminability.
\begin{table*}[!t]
\renewcommand\arraystretch{1.2}
\caption{Sensitivity of each class on the Fundus-iSee and MIMIC-CXR datasets.}
\begin{center}
    \begin{tabular}{|l|c|c|c|c|c|c|c|c|c|}
    \hline 
    \multirow{2}{*}{\textbf{Methods}} & \multicolumn{5}{c|}{\textbf{Fundus-iSee Dataset}} & \multicolumn{4}{c|}{\textbf{MIMIC-CXR Dataset}}\\
    \cline{2-10}
     & \textbf{Normal} & \textbf{AMD} & \textbf{DR} & \textbf{Glaucoma} & \textbf{Myopia} & \textbf{Normal} & \textbf{Edema} & \textbf{Pneumonia} & \textbf{Fracture} \\
    \hline
    ATH-CE & 0.7745 & 0.2669 & 0.2107 & 0.2198 & 0.2739 & 0.7868 & 0.3220 & 0.6637 & 0.2188\\
    ATH-focal & 0.7747 & 0.2755 & 0.2055 & 0.2201 & 0.2843 & 0.8149 & \underline{0.3305} & \underline{0.6998} & 0.2562\\
    ATH-circle & 0.7902 & \underline{0.2855} & \underline{0.2140} & \underline{0.2288} & \underline{0.2906} & 0.7783 & 0.3059 &  0.6858 & \underline{0.2683}\\
    ATH-pairwise & \underline{0.8218} & 0.2594 & 0.1852 & 0.1881 & 0.2394 & \underline{0.8303} & 0.2806 & 0.6375 &  0.2186\\
    ATH-triplet & 0.7918 & 0.2794 & 0.2052 & 0.2081 & 0.2894 & 0.8018 & 0.3194 & 0.6852 & 0.2581\\
    \textbf{ATH(ours)} & \textbf{0.8375} & \textbf{0.4027} & \textbf{0.3132} & \textbf{0.3216} & \textbf{0.3955} & \textbf{0.8511} & \textbf{0.4552} & \textbf{0.7864} & \textbf{0.3728}\\
    \hline
    \end{tabular}
\end{center}\label{tb4}
\end{table*}

\begin{figure}[htb]
  \centering
  \includegraphics[width=\linewidth]{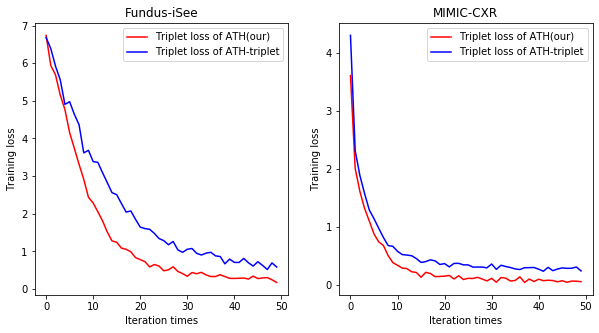}
  \caption{Qualitative results of training loss. The triplet loss value between our ATH (red line) and ATH-triplet (blue line) is compared on the Fundus-iSee (left plot) and MIMIC-CXR (right plot) datasets.}
  \label{fig6}
\end{figure}
By preserving the information of small samples and their class, we would like to demonstrate the cross-entropy loss can help to minimize intra-class similarity and maximize inter-class similarity. Although the circle loss and our triplet cross-entropy loss all combine the triplet loss and the cross-entropy loss, our triplet cross-entropy loss keeps both original forms to train on the triplet labels. The original form of the triplet loss punishes the similarity distance, the effect of which can thus be compared by observing the training loss value. We can easily extract the triplet loss value from the sum loss in our ATH and compare it to the triplet loss in the ATH-triplet. As shown in Fig.~\ref{fig6}, with the help of the cross-entropy loss, the triplet loss value of our ATH is lower than the ATH-triplet. According to Equation \ref{eq3}, a lower triplet loss value indicates the better hash code-discriminability. 

We further investigate the effectiveness of the weighting parameter \begin{math}r\end{math} and the length of hash codes \begin{math}k\end{math} of the triplet cross-entropy loss. As shown in Table \ref{tb5}, we observe the performances over hash codes with lengths of 12, 24, 36, and 48 by setting the weighting parameter \begin{math}r\end{math} as 0.3, 0.5, and 0.7, respectively. The best performance is achieved by setting \begin{math}r=0.5\end{math} and \begin{math}k=24\end{math} on the Fundus-iSee dataset, and \begin{math}r=0.5\end{math} and \begin{math}k=36\end{math} on the MIMIC-CXR dataset. Reasonably, \begin{math}r=0.5\end{math} refers to that half of the hash code lengths between dissimilar images should be different. With the hash codes lengthen, \begin{math}r\end{math} can be correspondingly set higher than the short hash codes, then the performance can correspondingly improve at the cost of storage and search efficiency. As a trade-off between performance and search cost, we set \begin{math}r=0.5\end{math} and \begin{math}k=36\end{math} in our experiments.

\begin{table*}[!t]
\renewcommand\arraystretch{1.2}
\caption{mAP over the varying $r$ and $k$ of our method on Fundus-iSee and MIMIC-CXR datasets.}
\begin{center}
    \begin{tabular}{|c|c|c|c|c|c|c|c|c|}
    \hline 
    \multirow{2}{*}{\textbf{Parameter}} & \multicolumn{4}{c|}{\textbf{Fundus-iSee Dataset}} & \multicolumn{4}{c|}{\textbf{MIMIC-CXR Dataset}}\\
    \cline{2-9}
     & $k=12$ & $k=24$ & $k=36$ & $k=48$ & $k=12$ & $k=24$ & $k=36$ & $k=48$\\
    \hline
    $r=0.3$ & 0.6593 & 0.7051 & 0.6714 & 0.6431 & 0.7562 & 0.7164 & 0.7363 & 0.7433 \\
    $r=0.5$ & 0.6470 & \textbf{0.7322} & 0.7220 & 0.6504 & 0.7354 & 0.7780 & \textbf{0.8260} & 0.7572 \\
    $r=0.7$ & 0.6345 & 0.6534 & 0.6214 & 0.6659 & 0.7296 & 0.7434 & 0.7645 & 0.7718 \\
    \hline
    \end{tabular}
\end{center}\label{tb5}
\end{table*}

At the last analysis of experiments, the efficiency of the proposed method will be discussed on four-folds by putting the MIMIC-CXR dataset as an example. 
\begin{itemize}
 \item[1)] \textbf{Feature computation time.} Based on the pre-trained ATH model with 36-bit hash codes, the feature extraction of the training set of 20,000 images can be completed in 28 seconds by using GPU. 
 \item[2)] \textbf{Retrieval time.} After feature extraction mapping into 36-bit hash codes, the index is built in 1 second by using Faiss. Then the retrieval of the test set of 20,000 images can be done in 1,346 $ms$ by returning top-10 most similar images. 
 \item[3)]  \textbf{Training time.} Training our ATH with an end-to-end manner takes 2,325 seconds by setting the iteration number as 50. 
 \item[4)] \textbf{Memory cost.} During model training with a batch size of 10, the memory-consuming is about 2,000 Mbps. The online search for the index also consumes about 2,000 Mbps. 
\end{itemize}
Compared to the state-of-the-art methods, including DPSH-pairwise, DRH-pairwise, DSH-triplet, and DBEN-triplet, the complexity of our ATH is slightly higher in training time and memory cost due to the added attention module and is fair in feature computation time and retrieval time. According to the above analysis of efficiency, our ATH can provide fair real-time responses with significantly improving the performance, compared to the state-of-the-art deep hashing methods.

\section{Conclusions}
\label{sec_con}
To enhance the ranking quality of case-based medical image retrieval, the proposed Attention-based Triplet Hashing network (ATH) is able to preserve classification, regions of interest (ROI), and small-sample information in the hashing space. We embed a spatial-attention module into the network to capture the ROI information. A novel triplet cross-entropy loss is proposed to preserve the classification information in the hash codes by punishing the similarity and classification losses simultaneously. Further, the triplet labels can fully utilize small samples to alleviate the imbalanced-sample problem to some extent. Experiments on two case-based medical datasets, including fundus images and chest X-rays, demonstrate that our ATH can obtain state-of-the-art performances in medical image retrieval. Further analysis confirms that the triplet cross-entropy loss can enhance classification performance and hash code-discriminability. Although our ATH achieves competitive performance, two promising directions for future research can be devoted to case-based medical image retrieval. First, we can integrate the triplet cross-entropy loss with regularizer to further differentiate images better for small samples, referring to existing hashing methods. Second, to focus on ROI information, we can design a new region-wise attention network that weights an attentive score of a region considering attentiveness.

\section*{Acknowledgments}
The authors would like to thank many members of the Intelligent Medical Imaging (iMED) group for the inspiring knowledge sharing, technical discussions, clinical background infusion. Their helping hands make this paper a reality. We also would like to thank our partner Xiaoai Clinic for the data support. This work was supported in part by Guangdong Provincial Key Laboratory of Brain-inspired Intelligent Computation (Grant No. 2020B121201001). 
%%Harvard
\bibliographystyle{model2-names.bst}\biboptions{authoryear}
\bibliography{mia}

\end{document}